\documentclass[sigconf]{acmart}
\AtBeginDocument{%
  \providecommand\BibTeX{{%
    \normalfont B\kern-0.5em{\scshape i\kern-0.25em b}\kern-0.8em\TeX}}}
    
\usepackage{float}
\usepackage{subcaption}

\copyrightyear{2021} 
\acmYear{2021} 
\setcopyright{acmlicensed}
\acmConference[CHI '21 Workshop SpaceCHI]{The 2021 ACM CHI Virtual Conference on Human Factors in Computing Systems}{May 8-13, 2021}{Yokohama, Japan (Online Virtual Conference)}
\acmBooktitle{CHI '21 workshop SpaceCHI: Human-Computer Interaction for Space Exploration}

\begin{document}

\title{Toward Designing Social Human-Robot Interactions for Deep Space Exploration}

\author{Huili Chen}
\affiliation{%
  \streetaddress{MIT Media Lab}
   \city{MIT Media Lab}}
\email{hchen25@media.mit.edu}

\author{Cynthia Breazeal}
\affiliation{%
   \city{MIT Media Lab}}
\email{cynthiab@media.mit.edu}

\begin{abstract}
In planning for future human space exploration, it is important to consider how to design for uplifting interpersonal communications and social dynamics among crew members. What if embodied social robots could help to improve the overall team interaction experience in space? On Earth, social robots have been shown effective in providing companionship, relieving stress and anxiety, fostering connection among people, enhancing team performance, and mediating conflicts in human groups. In this paper, we introduce a set of novel research questions exploring social human-robot interactions in long-duration space exploration missions.  
\end{abstract}


\maketitle

\section{Introduction}
\subsection{Human Factors in LDSE Missions}\label{human-factors-ldse}
NASA is actively planning for long-duration space exploration (LDSE) missions such as an anticipated manned mission to Mars in the 2030s~\cite{nasa-2014}. 
In planning for future LDSE missions, one major risk area pertaining to human factors concerns the ability of astronauts to adapt to the isolated, confined, and extreme (ICE) conditions~\cite{nasa_2020}. Living in ICE conditions for a long duration is highly stressful. The psycho-environmental factors of living and working in such ICE environments, as documented by studies of these ICE analogues and evidence from past spaceflights, include crowding, lack of privacy, social isolation, and sensory restriction~\cite{Palinkas1995EffectsOP,Raybeck1991-ProxemicsPM}. The sense of isolation will become further heightened with longer distances from Earth and communication lag with Earth~\cite{nasa_2020}. 

To ensure the success of a LDSE mission, adaptation is needed for both individual astronauts and the crew team as a whole. On the individual level, astronauts will need to cope with a variety of observed behavioral, physiological and psychological problems including anger, anxiety, interpersonal conflict, social withdrawal, sleep deprivation, decrease in group cohesion, and decrease in motivation~\cite{Flynn2005-OperationalAL}. 
On the team level, factors related to interpersonal communications and group dynamics among astronauts also decisively impact mission success. For example, astronauts will have to live in an ICE condition for the entirety of the mission, necessitating group living skills to combat potential interpersonal problems~\cite{Galarza1999-CriticalAP}. Their diverse cultural backgrounds may additionally impact coping in an ICE environment and interplanetary crew's behavior~\cite{TAFFORIN2017-CulturalEN}. Unlike shorter-duration space missions, LDSE missions require astronauts to have an unprecedented level of autonomy, leading to greater importance of interpersonal communication among crew members for mission success. The real-time support and interventions from human specialists on Earth (e.g., psychologist, doctor, conflict mediator) are reduced to minimal due to costly communication and natural delays in communication between space and Earth. All these human factor risks could lead to serious detrimental impacts on a LDSE mission if left unmitigated. To address them, socially interactive technologies such as social robots offer great opportunities of delivering effective interventions.

\subsection{Human-robot Interaction (HRI) in Space Exploration}
In the past decades, most research on HRI in space exploration focused on the engineering and cognitive aspects of space robotics. Space robots were often envisioned with advanced cognitive systems (e.g.,CARACaS) capable of model building, continuous planning/re-planning, self-diagnosis, and novel situation understanding~\cite{Huntsberger2010-EnvisioningCR}. These space robots could assist crew members with a variety of physical and cognitive tasks, but they did not necessarily support social and affective communications with crew members. In recent years, sociable space robots started to emerge, able to display and recognize social cues. An in-cabin flying robot named Astronaut Assistant Robot (AAR)~\cite{Liu2016-AttitudeCA} allows astronauts to use hand gestures to communicate with it face-to-face~\cite{Gao2017-StaticHG}. As the first free-floating, sphere-shaped interactive companion robot in space, Crew Interactive Mobile Companion (CIMON) designed by NASA and IBM~\cite{Murphy2018-CIMON} can display human-like facial expressions on its screen and respond to voice questions or directions without the need for a tablet or computer when assisting astronauts in their daily work. 

The potential benefits of social robots in space are also suggested by prior HRI work conducted on Earth, which found social robots effective in improving people's mental health and well-being~\cite{Scoglio2019-UseSR} as well as human group dynamics and team performance~\cite{Sebo2020RGT}. In LDSE missions, the needs of astronauts for social robots will very likely become heightened given the LDSE-related human factor risks listed in Section~\ref{human-factors-ldse}, thereby providing a greater motivation for more research on social HRI in space. In this paper, we focus on the opportunity of using social robots as a multimodal medium for ameliorating interpersonal communications and group dynamics of crew members in LDSE missions.


\section{Social HRI on Earth}
\subsection{Robot Social Roles in Human Groups}
Social robots can play a variety of roles ranging from highly knowledgeable agents to social-emotional companions across human group interaction contexts. When engaging with a human group as a highly knowledgeable agent and expert in a task area, a social robot, for example, can provide counseling to couples~\cite{Utami2019CUR}, offer direction guide in public space~\cite{Fraune2019HGP}, host group game activities~\cite{Zarkowski2019MTR}, and allocate turns to players~\cite{Claure2020MAB}. In contrast to an expert role, a robot can take a peer role that strategically displays its vulnerability or emotions with the goal of ameliorating human group dynamics. For example, a robot \textit{Kip1} was designed as a peripheral conversation companion that promotes non-aggressive conversation between people by expressing either curious interest or fear via its gesture cues~\cite{Hoffman2015-DesignEP}. Similarly, vulnerable statements made by a robot in a human group can positively impact the dynamics and collaboration among the group members (e.g., more equal conversation distribution and more positive group perception~\cite{Traeger2020VRP}). 

In addition to the expert and social-emotional companion roles, a robot can take a mediator role to directly resolve human team conflicts. For example, a social robot can improve people's interpersonal conflict resolution skills by flagging a conflict onset and offering prompts for conflict resolution~\cite{Shen2018SIS}. Mediation via the Telenoid robot was also found to produce more agreements and more integrative agreements among human teammates in comparison with both a screen mediator and a human mediator~\cite{Druckman2020-WhoBM}. The last social role category that a robot often takes in a human group is a moderator/supporter role. When taking this role, a robot could improve a human group's social dynamics~\cite{Short2016ModelingMM} and a dyadic human team's performance on a collaborative task by posing questions~\cite{Strohkorb2016ImprovingHC}. 

Overall, a robot's social role in a human group interaction distinctly shapes how humans interact with each other and how they perceive the robot.

\subsection{Robot Social Behaviors in Human Groups}
Both of a robot's verbal and nonverbal social behaviors can positively influence human groups when displayed appropriately, though in their unique ways. Specifically, a robot’s verbal communication can support its expressions of emotion~\cite{Leite2013TheIO,Correia2018GBE} and share informational content~\cite{Sabelli2016-RobovieMQ} with humans. It can also improve a variety of affective phenomena in human-human interactions such as trust building~\cite{Correia2018GBE}, group engagement~\cite{Matsuyama2015FGC}, psychological safety of outgroup team members~\cite{Sebo2020-InfluenceRV}, equality and positivity within conversational groups~\cite{Traeger2020VRP}, and the inclusion of human members within a team~\cite{Sebo2020GTH}. 

In addition to its verbal utterances, a robot’s nonverbal behaviors such as gestures~\cite{Liu2013-ItPP}, gaze~\cite{Mutlu2009FHR,Skantze2017-PredictingRP} navigation~\cite{Kidokoro2013-WillIB,Mavrogiannis2019-EffectsDR}, and physical orientation~\cite{vazquez2017TRA,Shiomi2010-LargerAP} can influence people's responses in a group interaction and their perception of the group~\cite{Sebo2020RGT}. For example, robots were designed, in some interactions~\cite{Tennent2019MPR,Fink2014WRB}, as peripheral or passive entities that shaped human group dynamics via implicit nonverbal behaviors without eliciting users' awareness. Often inspired by Ju's theory of implicit interactions~\cite{Ju2008TheDO}, this alternative design situates a robot as a passive entity in a group. For example, a robotic microphone that exhibits implicit engagement behaviors (e.g., nonverbal backchanneling) in a group problem solving context encouraged more active participation from passive human members and promoted more effective group collaboration performance~\cite{Tennent2019MPR}. As shown above, a robot's social influence on human groups can be achieved through both its use of speech and nonverbal social cues.  

Nevertheless, an excessive or inappropriate display of social cues may distract or interrupt humans in some circumstances~\cite{Kennedy2015TheRW,Yadollahi2018WDG}. For example, the frequency and length of a robot's speech affected humans' interaction outcomes more significantly than its content due to the disturbing effects of the inappropriate speech display~\cite{Jung2015-UsingRM,Short2017RMC}. Penalizing high frequency of speech as the “communication cost” can yield more effective human-robot communication~\cite{Vaibhav2020-DecisionBC}. For nonverbal behaviors, a robot's congruent display of social cues, e.g., contingent gaze and pointing gestures, can elicit greater human participation in an interaction~\cite{Lohan2012-ContingencySL}, while the asynchrony of social cues, e.g. robot head-gazes and pointing gestures made in different directions, may lead to interference effects and slow down human processing time~\cite{Langton2000-MutualIG,Langton2000YouMS}.

Hence, understanding the relations between the display of robot social behaviors and interaction contexts is crucial for designing successful multi-party HRI. 



\subsection{Human Trust in Robots}
Trust is a result of dynamic interactions~\cite{Mayer1995-IntegrativeMO}. Successful interactions will lead to feelings of security, trust and optimism, while failed interactions may result in human's unsecured feelings or mistrust. Much empirical evidence shows trust essential for successful human-human interactions~\cite{McAllister1995-AffectCT}. In human-robot interactions, humans' trust in robots also plays a critical role such as influencing their willingness to accept information from robots~\cite{Hancock2011-MetaFA} and to cooperate with robots~\cite{Freedy2007-MeasurementTH}. 
A person's trust in a robot is often intertwined with the robot's task performance, display of nonverbal behaviors, and interaction personalization. A single error of the robot can impact humans’ trust of it, especially in critical situations~\cite{Robinette2017-EffectRP}. If a robot displays nonverbal signals that humans often exhibit to indicate distrust, the robot would also be perceived as less trustworthy~\cite{Desteno2012-DetectingTN}. In the context of workplace-based long-term HRI, a personalized human-robot discussion was found to increase the person's rapport and cooperation with the robot as compared with a social but not personalized discussion~\cite{Lee2012PersonalizationIH}. Humans' trust in robots, given its importance in HRIs, is often used as an evaluation measure on robot decision making~\cite{Chen2018-PlanningTH} and human-robot team effectiveness~\cite{Freedy2007-MeasurementTH} across contexts and tasks. 

\section{Future Social HRI in Space Exploration}
Despite extensive HRI research conducted on Earth, we barely know to what extent these prior findings can be readily applied to the deep space exploration context. Contextualizing the HRI research in LDSE missions would unlock the full potential of using social robots to foster positive interpersonal communications and social dynamics among crew members in future space exploration. Hence, we propose a set of key design questions \textbf{(DQ)} pertaining to social HRI in space.

\subsection{Robot Social Role Design}
When interacting with a human group, social robots have been taking a variety of roles from resource providers to listeners, each of which offers unique benefits contingent on the interaction context. However, barely any prior work focused on the design and impact of robot roles on the group dynamics and processes of space crew teams, posing a new urgency to investigate the nuanced challenges for social HRI in space. For example, a robot mediator has been empirically shown to resolve people's interpersonal conflicts more effectively than a screen mediator or a human mediator~\cite{Druckman2020-WhoBM}, but the design considerations for a robot mediation in astronaut-astronaut interactions remain unexplored. Understanding astronauts' perception, acceptance and needs of social robots in different LDSE-related interaction contexts would help design for more astronaut-centered HRI. Therefore, we propose the first two research questions as follows: 
\begin{itemize}
    \item \textbf{DQ1:} What robot interactions would astronauts perceive to be socially, cognitively and affectively beneficial to the crew team in a LDSE mission?
    \item  \textbf{DQ2:} How would individual characteristics and cultural backgrounds of astronauts affect their perception, acceptance and preference of social HRI in space? 
\end{itemize}

In LDSE missions, crew members engage in a variety of group interactions ranging from highly-critical team cooperation and urgent problem-solving meetings to leisure and recreational activities. In these activities, the crew team's social dynamics may vary as the social roles of the members are adaptive to the context, e.g., superior-subordinate communication and peer interaction. The way a social robot engages with the team should thus be contingent on the team's social dynamics. Designing a diverse set of robot social roles customized to different group interaction contexts could potentially promote the overall positive astronaut-robot interaction experience in space. We summarize this research topic as follows: 
\begin{itemize}
    \item \textbf{DQ3:} What robot role(s) could be designed for each crew team interaction context in a LDSE mission?
\end{itemize}

Since all group interactions take place in an ICE environment with limited physical space and resources for a long duration, different robot roles designed for specific group contexts, e.g., coach, peer and mediator, may likely have to share a robot hardware embodiment rather than each owning a different physical embodiment. This constraint on robot hardware resources further poses additional design challenges pertaining to robot role and identity switching, as summarized below. 

\begin{figure}
\begin{subfigure}[h]{0.7\linewidth}
 \includegraphics[width=1.0\linewidth]{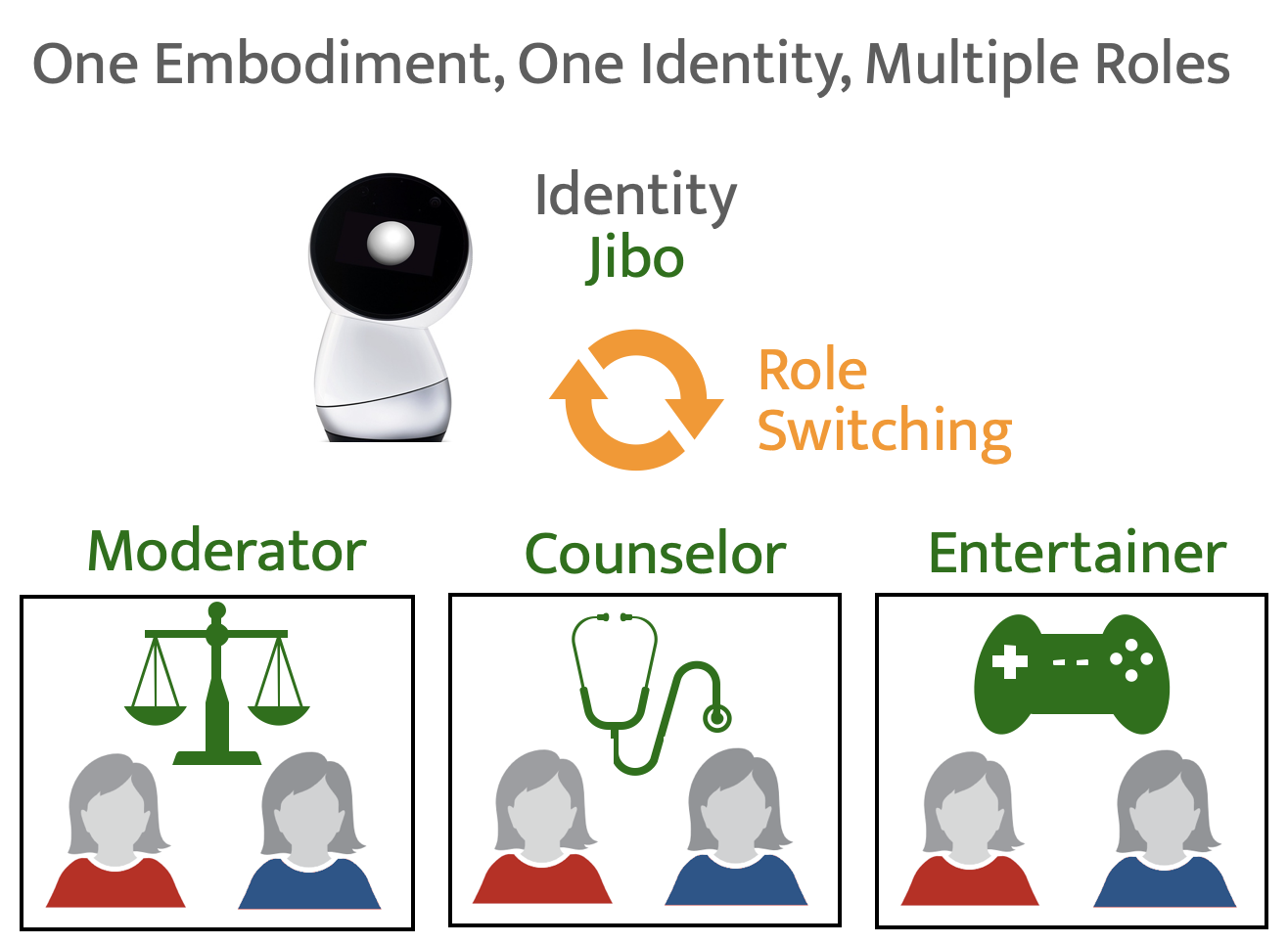}
 \caption{Robot Role Switching}
 \label{robot-role-switching}

\end{subfigure}
\hfill
\begin{subfigure}[h]{0.7\linewidth}
\includegraphics[width=1.00\linewidth]{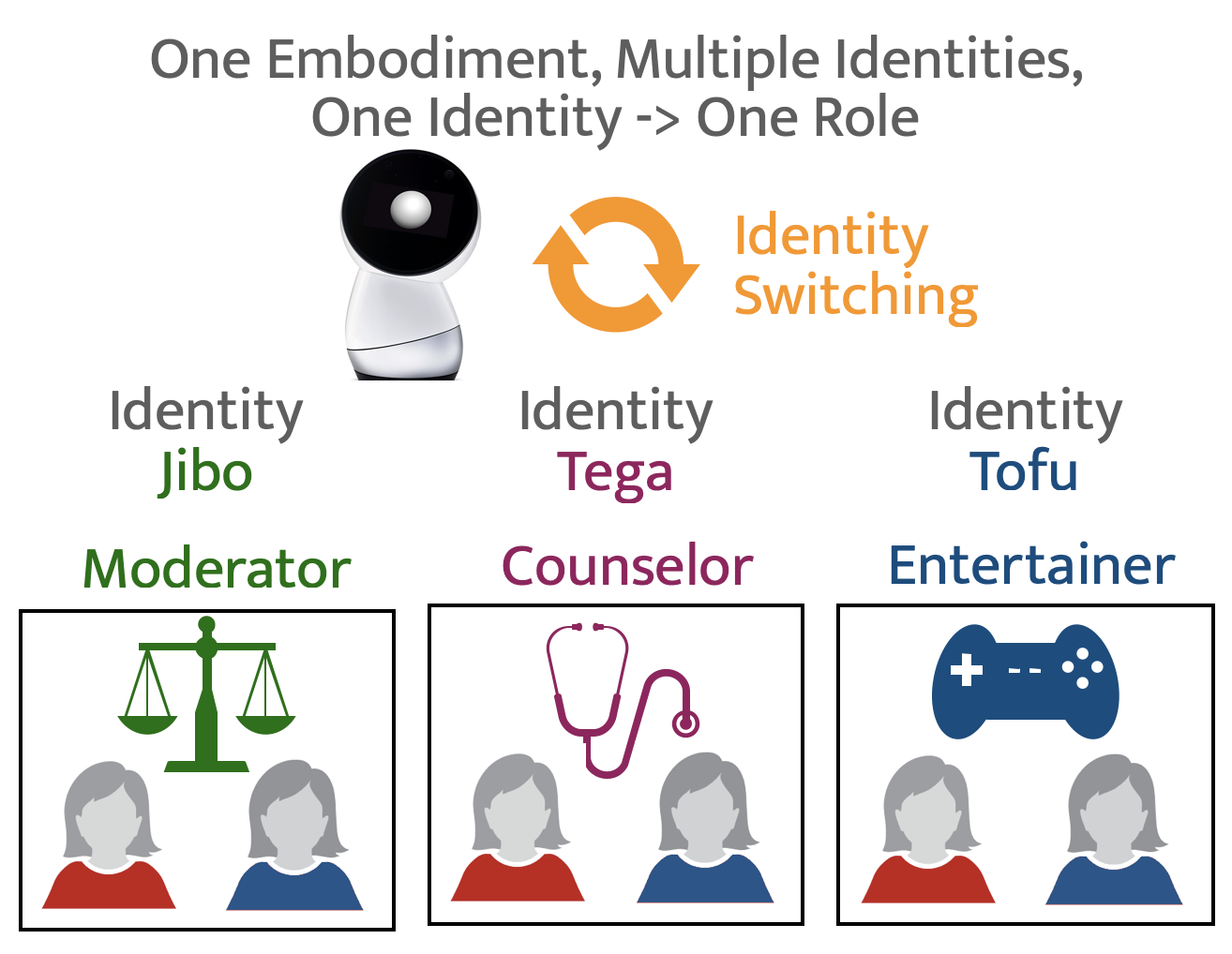}

\caption{Robot Identity Switching}
\label{robot-identity-switching}
\end{subfigure}
\caption{Two design approaches to address the constraint on robot hardware resources in LDSE missions.}
\label{robot-hardware-resource-constraint}
 \vspace{-5mm} 
\end{figure}

\begin{itemize}
   
    \item \textbf{DQ4:} Should a robot's social identity (e.g., name, memory and personality) remain consistent while its social role switches across interaction contexts, as illustrated in Figure~\ref{robot-role-switching}? Should its social identity switch across contexts so that each robot role is uniquely associated with a distinct robot identity, as illustrated in Figure~\ref{robot-identity-switching}?
     \item \textbf{DQ5:} How should a physical robot embodiment identify group contexts and dynamics, and proactively switch its social role or identity to effectively promote positive crew team experience?
    \item \textbf{DQ6:} How would the aforementioned role-switching and identity-switching approaches influence astronauts' perception and acceptance of robot interactions such as their trust in robots? 
\end{itemize}

\subsection{Robot Social Behavior Design}
Robots can positively influence a human group via diverse combinations of social cues, from actively utilizing animated speech to solely using nonverbal cues without eliciting human awareness. In order to be interpreted correctly and efficiently, robot social cues need be contingent on specific interactions as well as congruent with other social cues being utilized. Unlike the Earth environments where the most prior HRI studies were conducted, outer space environment imposes much more stressful challenges on human bodies and minds, resulting in a variety of expected physiological, psychological, cognitive and physical changes of astronauts. Investigating how astronauts encode and decode multimodal social cues when interacting with a robot in various physical and mental conditions would help design for space robots that can adapt their social-affective expression styles based on both physical environment and in-the-moment states of astronauts to improve human-robot and human-human communications. Thus, we summarize this design question on robot social cues as follows:

\begin{itemize}
    \item \textbf{DQ7:} How would different gravity conditions, e.g., microgravity, lunar and Martian gravity, impact astronauts' social, cognitive and affective reactions to a variety of robot social cues, e.g., gaze sharing, body motion and speech, in both group and single-person HRI contexts? What social cues would astronauts prefer to use for human-robot communications in different gravity conditions? 
\end{itemize}
 
Living in ICE conditions for a long duration exposes astronauts to long-term sensory restriction. Unaffected by ICE conditions, social robots hence have great potential of delivering high-quality consistent multi-sensory intervention to astronauts. This design question is summarized as follows:  

\begin{itemize}
     \item \textbf{DQ8:} How could different robot social cues and behaviors be designed to provide positive multi-sensory stimulation to the crew team as well as enhance multimodal social-affective communications among crew members, e.g., rapport, social touch, and social reciprocity?
\end{itemize}

\subsection{Robot Social Adaptation Design}
Human adaptability for deep space environment is crucial for mission success. In addition to traditional coping and resource strategies for astronauts' adaptation~\cite{Bartone2019-HumanAD}, a space robot's social adaptability to astronauts could also facilitate their adaptation to deep space with the long-term goal of maintaining their health and productivity. Multiple social-affective signal modalities of astronauts could be leveraged to develop a robot social adaptation system, including their physiological signals (e.g., blood pressure, sleep patterns), physical conditions (e.g., malnutrition), as well as their psychological and cognitive states (e.g., depressive mood, anxiety, loneliness), alongside the robot's audiovisual observation of the naturalistic crew interactions. For example, if a robot can monitor each astronaut's sleep quality, it can adapt its social role and social cues to minimize the negative effect of astronauts' sleep deprivation on team decision making and other cognitive processes. The access to a more diverse set of astronauts' social-affective signal modalities would enable a robot's more strategic and timely adaptation. However, a robot's overaggressive data-driven adaptation may increase the risk of exacerbating astronauts' perceived lack of privacy in an ICE environment and weakening their interpersonal trust in the robot and the team, potentially resulting in lower overall crew team dynamics and productivity. Investigating the optimal robot adaptation in LDSE missions requires an astronaut-centered design approach. Therefore, we propose two astronaut-centered design questions as follows: 

\begin{itemize}
    \item \textbf{DQ9:} What social-affective signal modalities of astronauts could a robot have access to for its long-term social adaptation?
    \item \textbf{DQ10:} How to design a robot's social adaptation system that can maintain and even foster crew members' perceived privacy and interpersonal trust in the robot and other members in LDSE missions? 
\end{itemize}

\section{Conclusion}
This paper introduces novel opportunities for social human-robot interactions in deep space exploration. We believe that exploring the design questions listed in the paper would help shed light on how social robotics could be potentially used to promote interpersonal communications and facilitate group interactions among crew members in long-duration space exploration missions.

\bibliographystyle{ACM-Reference-Format}
\bibliography{paper-bib}
\end{document}